\documentclass[11pt,a4paper]{article}
\pdfoutput=1

\usepackage[hyperref]{acl2021}
\usepackage{times}
\usepackage{latexsym}

\usepackage{microtype}

\usepackage{tikz}
\usepackage{tipa}
\usetikzlibrary{positioning,arrows,arrows.meta,shapes.misc,calc,backgrounds}
\usepackage{listings}
\usepackage{color}
\usepackage{booktabs}
\usepackage{array}
\usepackage{amsmath,amssymb}
\usepackage{multirow}
\usepackage{caption}
\usepackage{subcaption}
\usepackage{inconsolata}
\usepackage[inline]{enumitem}
\usepackage{pifont}
\usepackage{soul}
\usepackage[capitalise,noabbrev]{cleveref}

\definecolor{mygreen}{RGB}{0,153,0}

\lstnewenvironment{snippet}{
\lstset{ 
  backgroundcolor=\color{white},
  basicstyle=\footnotesize\ttfamily\bfseries,
  breakatwhitespace=false,
  breaklines=false,
  escapeinside={\%*}{*)},
  extendedchars=true,
  keepspaces=true, 
  morekeywords={Else,If}, 
  rulecolor=\color{black},
  showspaces=false,
  showstringspaces=false,
  framextopmargin=0pt,
  framexbottommargin=0pt,
  showtabs=false,
  stepnumber=2,
  tabsize=1,
  title=\lstname
  }
}{}

\lstnewenvironment{programs}{
\lstset{ 
  backgroundcolor=\color{white},
  basicstyle=\footnotesize\ttfamily,%
  breakatwhitespace=false,
  breaklines=false,
  escapeinside={\%*}{*)},
  extendedchars=true,
  keepspaces=true, 
  morekeywords={Else,IfThen,Map}, 
  rulecolor=\color{black},
  showspaces=false,
  showstringspaces=false,
  showtabs=false,
  stepnumber=2,
  tabsize=1,
  title=\lstname
  }
}{}

\aclfinalcopy %

\title{Sample-efficient Linguistic Generalizations through Program Synthesis: Experiments with Phonology Problems}

\author{Saujas Vaduguru$^1$\;\;\;Aalok Sathe$^2$\;\;\;Monojit Choudhury$^3$\;\;\;Dipti Misra Sharma$^1$ \\
  ${}^1$ IIIT Hyderabad \hspace*{0.3in}
  ${}^2$ MIT BCS\thanks{\ \ Work done while at the University of Richmond} \hspace*{0.3in} \\
  ${}^3$ Microsoft Research India \\
  ${}^1$ {\tt \{saujas.vaduguru@research.,dipti@\}iiit.ac.in} \;\;\\ ${}^2$ {\tt aalok.sathe@\{mit.edu, richmond.edu\}} \;\;\\ ${}^3$ {\tt monojitc@microsoft.com} \\}

\date{}

\begin{document}
\maketitle
\begin{abstract}
Neural models excel at extracting statistical patterns from large amounts of data, but struggle to learn patterns or reason about language from only a few examples. In this paper, we ask: Can we learn explicit rules that generalize well from only a few examples? We explore this question using program synthesis. We develop a synthesis model to learn phonology rules as programs in a domain-specific language. We test the ability of our models to generalize from few training examples using our new dataset of problems from the Linguistics Olympiad, a challenging set of tasks that require strong linguistic reasoning ability. In addition to being highly sample-efficient, our approach generates human-readable programs, and allows control over the generalizability of the learnt programs. 
\end{abstract}

\section{Introduction}
In the last few years, the application of deep neural models has allowed rapid progress in NLP. Tasks in phonology and morphology have been no exception to this, with neural encoder-decoder models achieving strong results in recent shared tasks in phonology \cite{gorman-etal-2020-sigmorphon} and morphology \cite{vylomova-etal-2020-sigmorphon}. However, the neural models that perform well on these tasks make use of hundreds, if not thousands of training examples for each language. Additionally, the patterns that neural models identify are not interpretable. In this paper, we explore the problem of learning interpretable phonological and morphological rules from only a small number of examples, a task that humans are able to perform.

Consider the example of verb forms in the language Mandar presented in \cref{tab:sampleprob}. How would a neural model tasked with filling the two blank cells do? The data comes from a language that is not represented in large-scale text datasets that could allow the model to harness pretraining, and the number of samples presented here is likely not sufficient for the neural model to learn the task.

However, a human would fare much better at this task even if they didn't know Mandar. Identifying rules and patterns in a different language is a principal concern of a descriptive linguist \cite{brown2010concise}. Even people who aren't trained in linguistics would be able to solve such a task, as evidenced by contestants in the Linguistics Olympiads\footnote{\url{https://www.ioling.org/}}, and general-audience puzzle books \cite{bellos2020language}. In addition to being able to solve the task, humans would be able to express their solution explicitly in terms of rules, that is to say, a \textit{program} that maps inputs to outputs.

\begin{table}[t]
    \centering
    \begin{tabular}{cc}
    \toprule
        \textbf{to \textit{V}} & \textbf{to be \textit{V}ed} \\
    \midrule
        \textipa{mappasuN} & \textipa{dipasuN} \\
        \textipa{mattunu} & \textipa{ditunu} \\
        ? & \textipa{ditimbe} \\
        ? & \textipa{dipande} \\
    \bottomrule
    \end{tabular}
    \caption{Verb forms in Mandar \cite{mandar_problem}}
    \label{tab:sampleprob}
\end{table}

\textit{Program synthesis} \cite{ProgramSynthesisNow} is a method that can be used to learn programs that map an input to an output in a \textit{domain-specific language} (DSL). It has been shown to be a highly sample-efficient technique to learn interpretable rules by specifying the assumptions of the task in the DSL \cite{FlashFill}. 

This raises the questions
\begin{enumerate*}[label={(\roman*)}]
    \item Can program synthesis be used to learn linguistic rules from only a few examples?
    \item If so, what kind of rules can be learnt?
    \item What kind of operations need to explicitly be defined in the DSL to allow it to model linguistic rules?
    \item What knowledge must be implicitly provided with these operations to allow the model to choose rules that generalize well?
\end{enumerate*}

\begin{table*}[t]
    \centering
    \begin{small}
    \begin{tabular}{p{1.8in}p{3.7in}}
    \toprule
        \textbf{Predicate} & \\
    \midrule
        $\texttt{IsToken(}w, s, i\texttt{)}$ & Is $x$ equal to the token $s$? This allows us to evaluate matches with specific tokens. \\ \midrule
        $\texttt{Is(}w, f, i\texttt{)}$ & Is $f$ true for $x$? This allows us to generalize beyond single tokens and use features that apply to multiple tokens. \\ \midrule
        $\texttt{TransformationApplied(}w, t, i\texttt{)}$ & Has the transformation $t$ has been applied to $x$ in a previous pass? This allows us to reference previous passes in learning rules for the current pass. \\ \midrule
        $\texttt{Not(}p\texttt{)}$ & Negates the predicate $p$. \\
    \bottomrule
    \end{tabular}
    \end{small}
    \caption{Predicates that are used for synthesis. The predicates are applied to a token $x$ that is at an offset $i$ from the current token in the word $w$. The offset may be positive to refer to tokens after the current token, zero to refer to the current token, or negative to refer to tokens before the current token.}
    \label{tab:predicates}
\end{table*}

In this work, we use program synthesis to learn phonological rules for solving Linguistics Olympiad problems, where only the minimal number of examples necessary to generalize are given \cite{sahin-etal-2020-puzzling}. 
We present a program synthesis model and a DSL for learning phonological rules, and curate a set of Linguistics Olympiad problems for evaluation. 

We perform experiments and comparisons to baselines, and find that program synthesis does significantly better than our baseline approaches. We also present some observations about the ability of our system to find rules that generalize well, and discuss examples of where it fails.

\section{Program synthesis}

Program synthesis is ``the task of automatically finding programs from the underlying programming language that satisfy (user) intent expressed in some form of constraints'' \cite{ProgramSynthesisNow}. This method allows us to specify domain-specific assumptions as a language, and use generic synthesis approches like \textsf{FlashMeta} \cite{flashmeta} to synthesize programs.

The ability to explicitly encode domain-specific assumptions gives program synthesis broad applicability to various tasks. In this paper, we explore applying it to the task of learning phonological rules. Whereas previous work on rule-learning has focused on learning rules of a specific type \cite{brill-1992-simple,johnson-1984-discovery}, the DSL in program synthesis allows learning rules of different types, and in different rule formalisms.

In this work, we explore learning rules similar to rewrite rules \cite{chomsky1968sound} that are used extensively to describe phonology. Sequences of rules are learnt using a noisy disjunctive synthesis algorithm \textsf{NDSyn} \cite{iyer2019synthesis} extended to learn \textit{stateful multi-pass rules} \cite{prolinguist}.

\subsection{Phonological rules as programs}

The synthesis task we solve is to learn a program in a domain-specific language (DSL) for string transduction, that is, to transform a given sequence of input tokens
$i \in \mathcal{I}^*$ to a sequence of output tokens $o \in \mathcal{O}^*$, where $\mathcal{I}$ is the set of input tokens, and $\mathcal{O}$ is the set of output tokens. Each token is a symbol accompanied by a feature set, a set of key-value pairs that maps feature names to boolean values. 

We learn programs for token-level examples, which transform an input token in its context to output tokens. The program is a sequence of rules which are applied to each token in an input string to produce the output string. The rules learnt are similar to rewrite rules, of the form
\begin{align*}
    \phi_{-l}\cdots\phi_{-2}\phi_{-1}X\phi_1\phi_2\cdots\phi_r &\rightarrow T
\end{align*}
where 
\begin{enumerate*}[label={(\roman*)}]
    \item $X: \mathcal{I} \rightarrow \mathbb{B}$ is a boolean predicate that determines input tokens to which the rule is applied
    \item $\phi_i: \mathcal{I} \rightarrow \mathbb{B}$ is a boolean predicate applied to the $i$\textsuperscript{th} character relative to $X$, and the predicates $\phi$ collectively determine the context in which the rule is applied
    \item $T: \mathcal{I} \rightarrow \mathcal{O}^*$ is a function that maps an input token to a sequence of output tokens.
\end{enumerate*}

$X$ and $\phi$ belong to a set of predicates $\mathcal{P}$, and $T$ is a function belonging to a set of transformation functions $\mathcal{T}$. $\mathcal{P}$ and $\mathcal{T}$ are specified by the DSL.

We allow the model to synthesize programs that apply multiple rules to a single token by synthesizing rules in passes and maintaining state from one pass to the next. This allows the system to learn stateful multi-pass rules \cite{prolinguist}.

\subsection{Domain-specific language}

\begin{table*}[t]
    \centering
    \begin{small}
    \begin{tabular}{p{1.5in}p{4in}}
    \toprule
        \textbf{Transformation} & \\
    \midrule
        $\texttt{ReplaceBy(}x, s_1, s_2\texttt{)}$ & If $x$ is $s_1$, it is replaced with $s_2$. This allows the system to learn conditional substitutions. \\ \midrule
        $\texttt{ReplaceAnyBy(}x, s\texttt{)}$ & $x$ is replaced with $s$. This allows the system to learn unconditional substitutions. \\ \midrule
        $\texttt{Insert(}x, S\texttt{)}$ & This inserts a sequence of tokens $S$ after $x$ at the end of the pass. It allows for the insertion of variable-length affixes. \\ \midrule
        $\texttt{Delete(}x\texttt{)}$ & This deletes $x$ from the word at the end of the pass. \\ \midrule
        $\texttt{CopyReplace(}x, i\texttt{)}$ & \multirow{4}{4in}{These are analogues of the \texttt{ReplaceBy} and \texttt{Insert} transformations where the token which is added is the same as the token at an offset $i$ from $x$. They allow the system to learn phonological transformations such as assimilation and gemination.} \\
        $\texttt{CopyInsert(}x, i\texttt{)}$ & \\ 
        & \\
        & \\\midrule
        $\texttt{Identity(}x\texttt{)}$ & This returns $x$ unchanged. It allows the system where a transformation applies under certain conditions, but does not under others. \\ 
    \bottomrule
    \end{tabular}
    \end{small}
    \caption{Transformations that are used for synthesis. The transformations are applied to a token $x$ in the word $w$. The offset $i$ for the \texttt{Copy} transformations may be positive to refer to tokens after the current token, zero to refer to the current token, or negative to refer to tokens before the current token.}
    \label{tab:transforms}
\end{table*}

The domain-specific language (DSL) is the declarative language which defines the allowable string transformation operations.
The DSL is defined by a set of operators, a grammar which determines how they can be combined, and a semantics which determines what each operator does. 
By defining operators to capture domain-specific phenomena, we can reduce the space of programs to be searched to include those programs that capture distinctions relevant to the domain. This allows us to explicitly encode knowledge of the domain into the system.

Operators in the DSL also have a score associated with each operator that allows for setting domain-specific preferences for certain kinds of programs. We can combine scores for each operator in a program to compute a ranking score that we can use to identify the most preferred program among candidates. The ranking score can capture implicit preferences like shorter programs, more/less general programs, certain classes of transformations, etc.

The DSL defines the predicates $\mathcal{P}$ and set of transformations $\mathcal{T}$ that can be applied to a particular token. The predicates and transformations in the DSL we use, along with the description of their semantics, can be found in \cref{tab:predicates,tab:transforms}.

\begin{figure}
    \centering
    \vspace{-0.73cm}
    \begin{programs}
output := Map(disjunction, input_tokens)
disjunction := Else(rule, disjunction)
rule := transformation 
        | IfThen(predicate, rule);
    \end{programs}
    \vspace{-0.5cm}
    \caption{\texttt{IfThen-Else} statements in the DSL}
    \label{fig:ifthenelse}
\end{figure}

Sequences of rules are learnt as disjunctions of \texttt{IfThen} operators, and are applied to each token of the input using a \texttt{Map} operator (\cref{fig:ifthenelse}). The conjunction of predicates $X$ and $\phi$ that define the context are learnt by nesting \texttt{IfThen} operators.

A \texttt{transformation} produces an token that is tagged with the transformation that is applied. This allows for maintaining state across passes.

The operators in our DSL are quite generic and can be applied to other string transformations as well. In addition to designing our DSL for string transformation tasks, we allow for phonological information to be specified as features, which are a set of key-value pairs that map attributes to boolean values. While we restrict our investigation to features based only on the symbols in the input, more complex features based on meaning and linguistic categories can be provided to a system that works on learning rules for more complex domains like morphology or syntax. We leave this investigation for future work.

\subsection{Synthesis algorithm}

\begin{figure*}[t]
    \centering
    \resizebox{\textwidth}{!}{\begin{tikzpicture}
    \newcommand{\myarrow}{{Latex[length=3mm, width=2mm]}}
    \node[text width=2.5cm, align=center] (1) [draw, rounded corners=10pt] {\textsf{\textbf{Words}}\\
    \textipa{k\ae t} $\rightarrow$ \textipa{k\ae ts} \\
    \textipa{dOg} $\rightarrow$ \textipa{dOgz}};
    
    \node[text width=2.5cm, align=center] (2) [draw, rounded corners=10pt, right=1.5cm of 1] {\textsf{\textbf{Token-level examples}}\\
    \textipa{k} $\rightarrow$ \textipa{k} \\
    \textipa{\ae} $\rightarrow$ \textipa{\ae} \\
    \textipa{t} $\rightarrow$ \textipa{t} \\
    \underline{\ \ } $\rightarrow$ \textipa{s} \\
    \textipa{d} $\rightarrow$ \textipa{d}\\
    \textipa{O} $\rightarrow$ \textipa{O}\\
    \textipa{g} $\rightarrow$ \textipa{g}\\
    \underline{\ \ } $\rightarrow$ \textipa{z}\\};

    \node[text width=6cm, align=center] (4) [draw, rounded corners=10pt, right=1.25cm of 2] {\textsf{\textbf{Candidates}}
    \vspace{-0.5cm}
\begin{programs}
 #1. IfThen(IsToken(w,"$",1), 
      IfThen(Is(w,"voice",0), 
       Insert(x,"z")))
 #2. IfThen(IsToken(w,"t",0),
      IfThen(IsToken(w,"$",1),
       Insert(x,"s")))
 #3. IfThen(IsToken(w,"$",1), 
      Insert(x,"s")),
 #4. IfThen(IsToken(w,"g",0),
      IfThen(IsToken(w,"$",1),
       Insert(x,"z")))
 #5. IfThen(
       Not(IsToken(w,"$",1)),
      Identity(x))
    \end{programs}
    \vspace{-0.5cm}
    };
    
    \node[text width=2.35cm, align=center] (6) [draw, rounded corners=10pt, right=1.75cm of 4] {\textsf{\textbf{Rules}}
    \vspace{-0.5cm}
\begin{programs}
 Else(#1,
  Else(#3,
   Else(#5)
   )
 )
\end{programs}
    \vspace{-0.5cm}
    };
    
    \node[text width=2cm,align=center] (5) [below=.5cm of 6] {\textsf{Program}};
    
    \node[text width=2cm,align=center] (0) [left=.5cm of 1] {\textsf{Input examples}};
    
    \draw[-\myarrow, thick] (1.east) to node[above]{\textsf{align}} (2.west);
    \draw[-\myarrow, thick] (0.east) to node[above]{} (1.west);
    \draw[-\myarrow, thick] (6.south) to node[above]{} (5.north);
    \draw[-\myarrow, thick] (2.east) to node[above]{\textsf{FM}} (4.west);
    \draw[-\myarrow, thick] (4.east) to node[above]{\textsf{NDSyn}} (6.west);
    \draw[-\myarrow, thick, rounded corners=12pt] (6) -- +(0,3.5cm) -- ($ (1) +(0,3.5cm) $) node[midway, above] () {\textsf{multi-pass}} -- (1);
\end{tikzpicture}}
    \caption{An illustration of the synthesis algorithm. \textsf{FM} is \textsf{FlashMeta}, which synthesizes rules which are combined into a disjunction of rules by \textsf{NDSyn}. Here, rule \texttt{\#1} is chosen over \texttt{\#4} since it uses the more general concept of the \texttt{voice} feature as opposed to a specific token, and thus has a higher ranking score.}
    \label{fig:multipassalgo}
\end{figure*}
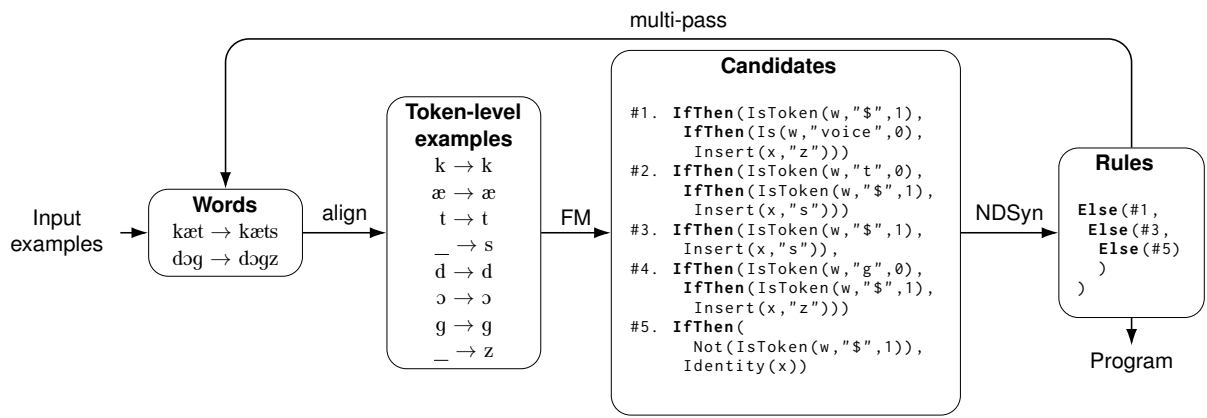

We use an extension \cite{prolinguist} of the \textsf{NDSyn} algorithm \cite{iyer2019synthesis} that can synthesize stateful multi-pass rules.
\citet{iyer2019synthesis} describe an algorithm for selecting disjunctions of rules, and use the \textsf{FlashMeta} algorithm as the rule synthesis component. \citet{prolinguist} extend the approach proposed by \citet{iyer2019synthesis} for disjunctive synthesis to the task of grapheme-to-phoneme (G2P) conversion in Hindi and Tamil. They propose the idea of learning transformations on token aligned examples, and use language-specific predicates and transformations to learn rules for G2P conversion. We use a similar approach, and use a different set of predicates and transformations that are language-agnostic.
\cref{fig:multipassalgo} sketches the working of the algorithm.

\begin{figure*}
    \centering
    \resizebox{\textwidth}{!}{\begin{tikzpicture}
\newcommand{\myarrow}{{Latex[length=3mm, width=2mm]}}
\node[text width=2cm, align=center] (1) [draw, rounded corners=5pt,fill=blue!20] {\textipa{a\underline{b}c} $\to$ \textipa{d}};

\node[text width=5cm, align=center] (2) [draw, rounded corners=5pt,fill=orange!20, below=0.75cm of 1] {\texttt{Inverse Semantics}\textsubscript{\texttt{IfThen}}};

\node[text width=2cm, align=center] (3) [draw, rounded corners=5pt,fill=blue!20,below left=0.75cm and -0.5cm of 2] {\textipa{a\underline{b}c} $\to$ \texttt{True}};

\node[text width=5cm, align=center] (4) [draw, rounded corners=5pt,fill=orange!20, below=0.75cm of 3] {\texttt{Inverse Semantics}\textsubscript{\texttt{IsToken}}};

\node[text width=1.75cm, align=center] (5) [draw, rounded corners=5pt,fill=blue!20,below left=0.75cm and -0.5cm of 4] {\textipa{a\underline{b}c} $\to$ \textipa{a}};

\node[text width=1.75cm, align=center] (6) [draw, rounded corners=5pt,fill=blue!20,below right=0.75cm and -0.5cm of 4] {\textipa{a\underline{b}c} $\to -1$};

\node[text width=1.75cm, align=center] (7) [draw, rounded corners=5pt,fill=blue!20,below=0.1cm of 5] {\textipa{a\underline{b}c} $\to$ \textipa{b}};

\node[text width=1.75cm, align=center] (8) [draw, rounded corners=5pt,fill=blue!20,below=0.1cm of 7] {\textipa{a\underline{b}c} $\to$ \textipa{c}};

\node[text width=1.75cm, align=center] (9) [draw, rounded corners=5pt,fill=blue!20,below=0.1cm of 6] {\textipa{a\underline{b}c} $\to 0$};

\node[text width=1.75cm, align=center] (10) [draw, rounded corners=5pt,fill=blue!20,below=0.1cm of 9] {\textipa{a\underline{b}c} $\to 1$};

\node[text width=3.6cm, align=center] (11) [draw, rounded corners=5pt,right=1cm of 9] {\vspace{-0.6cm}
\begin{programs}
 IsToken(w,"a",-1)
 IsToken(w,"b",0)
 IsToken(w,"c",1)
\end{programs}};

\node[text width=3cm, align=center] (12) [draw, dashed, rounded corners=5pt,fill=mygreen!20,below right=1.5cm and -0.5cm of 2] {\textit{Search for }\textsf{rule}};

\node[text width=4.1cm, align=center] (13) [draw, rounded corners=5pt,right=1cm of 12] {\vspace{-0.6cm}
\begin{programs}
 ReplaceBy(x,"b","d")
 ReplaceAnyBy(x,"d")
\end{programs}};

\node[text width=5.25cm, align=center] (14) [draw, rounded corners=5pt,right=2cm of 1] {\vspace{-0.6cm}
\begin{programs}
 IfThen(IsToken(w,"a",-1),
     ReplaceBy(x,"b","d"))
 IfThen(IsToken(w,"b",0),
     ReplaceAnyBy(x,"d"))
 IfThen(IsToken(w,"c",1),
     ReplaceAnyBy(x,"d"))
\end{programs}};

\begin{scope}[on background layer]
\draw[-\myarrow, thick] (1.south) to node[] {} (2.north);
\draw[-\myarrow, thick] ([xshift=-1cm, yshift=-0.3cm]2.center) to node[pos=0.75, text width=2.75cm, above]{\textsf{predicate}} (3.north);
\draw[-\myarrow, thick] ([xshift=1cm, yshift=-0.3cm]2.center) to node[pos=0.5, text width=2cm, auto]{\textsf{rule}} (12.north);
\draw[-\myarrow, thick] (3.south) to node[] {} (4.north);
\draw[-\myarrow, thick] ([xshift=-1cm, yshift=-0.3cm]4.center) to node[pos=0.75, text width=2cm, above]{\textsf{token}} (5.north);
\draw[-\myarrow, thick] ([xshift=1cm, yshift=-0.3cm]4.center) to node[pos=0.75, text width=2cm, auto]{\textsf{offset}} (6.north);

\draw[-{implies[scale=3]}, double, thick] ([xshift=0.1cm]9.east) to node[]{} ([xshift=-0.1cm]11.west);
\draw[-{implies[scale=3]}, double, thick] ([xshift=0.1cm]12.east) to node[]{} ([xshift=-0.1cm]13.west);
\draw[-{implies[scale=3]}, double, thick] ([xshift=0.1cm]1.east) to node[]{} ([xshift=-0.1cm]14.west);

\draw[-, dashed] (5.east) to node[]{} (6.west);
\draw[-, dashed] (7.east) to node[]{} (9.west);
\draw[-, dashed] (8.east) to node[]{} (10.west);
\end{scope}
\end{tikzpicture}}
    \caption{An illustration of the search performed by the \textsf{FlashMeta} algorithm. The \textcolor{blue}{blue} boxes show the specification that an operator must satisfy in terms of input-output examples, with the input token underlined in the context of the word. The \textcolor{orange}{\texttt{Inverse Semantics}} of an operator is a function that is used to infer the specification for each argument of the operator based on the semantics of the operator. This may be a single specification (as for \textsf{predicate}) or a disjunction of specifications (as for \textsf{token} and \textsf{offset}). The algorithm then recursively searches for programs to satisfy the specification for each argument, and combines the results of the search to obtain a program. The \textcolor{mygreen}{search} for the \textsf{rule} in an \texttt{IfThen} statement proceeds similarly to the search for a \textsf{predicate}. Examples of programs that are inferred from a specification are indicated with $\implies$. A dashed line between inferred specifications indicates that the specifications are inferred jointly.}
    \label{fig:flashmeta}
\end{figure*}
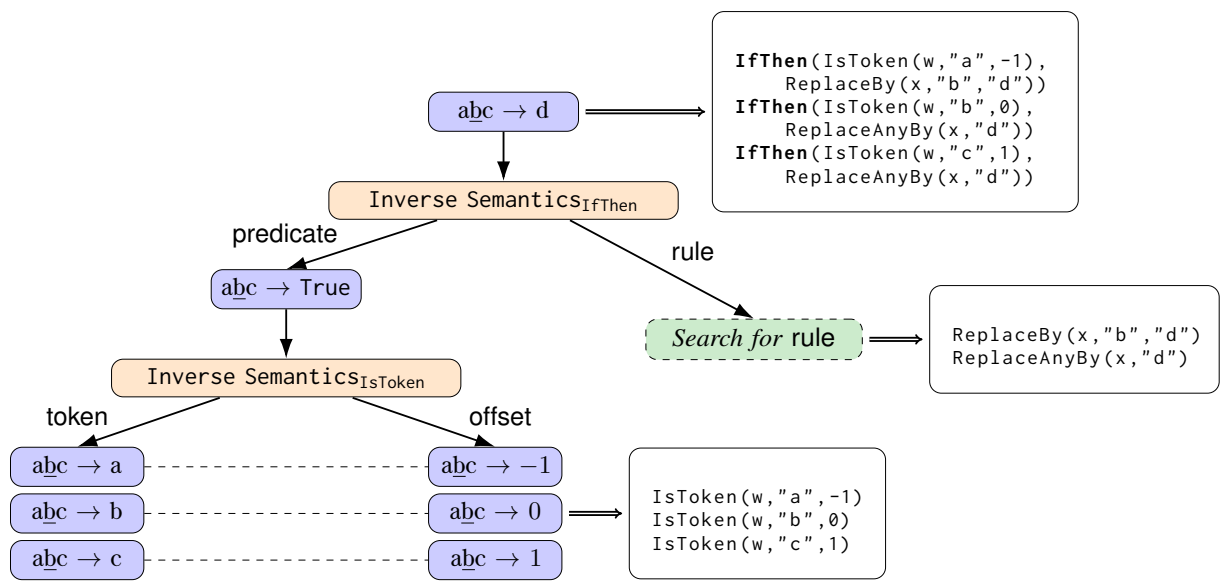

The \textsf{NDSyn} algorithm is an algorithm for learning disjunctions of rules, of the form shown in \cref{fig:ifthenelse}. Given a set of examples, it first generates a set of candidate rules using the \textsf{FlashMeta} synthesis algorithm \cite{flashmeta}. This algorithm searches for a program in the DSL that satisfies a set of examples by recursively breaking down the search problem into smaller sub-problems. Given an operator, and the input-output constraints it must satisfy, it infers constraints on each of the arguments to the operator, allowing it to recursively search for programs that satisfy these constraints on each of the arguments. For example, given the \texttt{Is} predicate and a set of examples where the predicate is true or false, the algorithm infers constraints on the arguments the token $s$ and offset $i$ such that the set of examples is satisfied. 
The working of \textsf{FlashMeta} is illustrated with an example in \cref{fig:flashmeta}.
We use the implementation of the \textsf{FlashMeta} algorithm available as part of the \textsc{prose}\footnote{\url{https://www.microsoft.com/en-us/research/group/prose/}} framework.

From the set of candidate rules, \textsf{NDSyn} selects a subset of rules with a high ranking score that correctly answers the most examples as well incorrectly answers the least\footnote{A rule will not produce any answer to examples that don't satisfy the context constraints of the rule.}. Additional details about the algorithm are provided in \cref{sec:ndsyn}.

The synthesis of multi-pass rules proceeds in passes. In each pass, a set of token-aligned examples is provided as input to the \textsf{NDSyn} algorithm. The resulting rules are then applied to all the examples, and those that are not solved are passed as the set of examples to \textsf{NDSyn} in the next pass. This proceeds until all the examples are solved, or for a maximum number of passes.

\section{Dataset} \label{sec:phonologyprobs}

To test the ability of our program synthesis system to learn linguistic rules from only a few examples, we require a task with a small number of training examples, and a number of test examples which measure how well the model generalises to unseen data. Additionally, to ensure a fair evaluation, the test examples should be chosen such that the samples in the training data provide sufficient evidence to correctly solve the test examples.

To this end, we use problems from the Linguistics Olympiad. The Linguistics Olympiad is an umbrella term describing contests for high school students across the globe. Students are tasked with solving linguistics problems---a genre of  composition that presents linguistic facts and phenomena in enigmatic form \cite{derzhanski_payne_2010}. These problems typically have 2 parts: the \textit{data} and the \textit{assignments}. 

The data consists of examples where the solver is presented with the application rules to some linguistic forms (words, phrases, sentences) and the forms derived by applying the rules to these forms. The data typically consists of 20-50 forms, the minimal number of examples required to infer the correct rules is presented \cite{sahin-etal-2020-puzzling}. 

The assignments provide other linguistic forms, and the solver is tasked with applying the rules inferred from the data to these forms. 
The forms in the assignments are carefully selected by the designer to test whether the solver has correctly inferred the rules, including making generalizations to unseen data. This allows us to see how much of the intended solution has been learnt by the solver by examining responses to the assignments.

The small number of training examples (data) tests the generalization ability and sample efficiency of the system, and presents a challenging learning problem for the system. The careful selection of test examples (assignment) lets us use them to measure how well the model learns these generalizations.

We present a dataset of 34 linguistics problems, collected from various publicly accessible sources. These problems are based on phonology, and some aspects of the morphology of languages, as well as the orthographic properties of languages. These problems are chosen such that the underlying rules depend only on the given word forms, and not on inherent properties of the word like grammatical gender or animacy.
The problems involve
\begin{enumerate*}[label={(\arabic*)}]
    \item inferring phonological rules in morphological inflection (\cref{tab:movima})
    \item inferring phonological changes between multiple related languages (\cref{tab:turkishtatar})
    \item converting between the orthographic form of a language and the corresponding phonological form (\cref{tab:micmac})
    \item marking the phonological stress on a given word (\cref{tab:aleut}).
\end{enumerate*}
We refer to each of these categories of problems as morphophonology, multilingual, transliteration, and stress respectively. We further describe the dataset in \cref{sec:dataappendix}\footnote{The dataset is available \href{https://github.com/saujasv/phonological-generalizations}{here}.}.

\begin{table*}[t]
    \small
    \centering
    \begin{subtable}{0.26\textwidth}
        \centering
        \begin{tabular}{ll}
        \toprule
        \textbf{base form} & \textbf{negative form} \\
        \midrule
        \textipa{joy} & \textipa{kas joya:ya’} \\
        \textipa{bi:law} & \textipa{kas bika’law} \\
        \textipa{tipoysu:da} & ?\\
        ? & \textipa{kas wurula:la’} \\
        \bottomrule
        \end{tabular}
        \caption{Movima negation} %
        \label{tab:movima}
    \end{subtable}
    \begin{subtable}{0.2\textwidth}
        \centering
        \begin{tabular}{ll}
        \toprule
            \textbf{Turkish} & \textbf{Tatar} \\
        \midrule
            \textipa{bandIr} & \textipa{mandIr} \\
            \textipa{yelken} & \textipa{cilk\"an} \\
            ? & \textipa{osta} \\
            \textipa{bilezik} & ? \\
        \bottomrule
        \end{tabular}
        \caption{Turkish and Tatar} %
        \label{tab:turkishtatar}
    \end{subtable}
    \begin{subtable}{0.26\textwidth}
        \centering
        \begin{tabular}{ll}
        \toprule
            \textbf{Listuguj} 
            & \textbf{Pronunciation} \\
        \midrule
            \textit{g’p’ta’q} & \textipa{g@b@da:x} \\
            \textit{epsaqtejg} & \textipa{epsaxteck} \\
            \textit{emtoqwatg} & ? \\
            ? & \textipa{@mtesk@m} \\
        \bottomrule
        \end{tabular}
        \caption{Micmac orthography}%
        \label{tab:micmac}
    \end{subtable}
    \begin{subtable}{0.24\textwidth}
        \centering
        \begin{tabular}{ll}
        \toprule
            \textbf{Aleut} 
            & \textbf{Stress} \\
        \midrule
            \textipa{tatul} & \texttt{01000} \\
            \textipa{n@tG@lqin} & \texttt{000010000} \\
            \textipa{sawat} & ? \\
            \textipa{qalpuqal} & \texttt{00001000} \\
        \bottomrule
        \end{tabular}
        \caption{Aleut stress}%
        \label{tab:aleut}
    \end{subtable}
    \caption{A few examples from different types of Linguistics Olympiad problems. `?' represents a cell in the table that is part of the test set.}
    \label{tab:exampleproblems}
\end{table*}

\subsection{Structure of the problems} \label{sec:probstructure}
Each problem is presented in the form of a matrix $M$. Each row of the matrix contains data pertaining to a single word/linguistic form, and each column contains the same form of different words, i.e., an inflectional or derivational paradigm, the word form in a particular language, the word in a particular script, or the stress values for each phoneme in a word. A test sample in this case is presented as a particular cell $M_{ij}$ in the table that has to be filled. The model has to use the data from other words in the same row ($M_{i:}$) and the words in the column ($M_{:j}$) to predict the form of the word in $M_{ij}$.

In addition to the data in the table, each problem contains some additional information about the symbols used to represent the words. This additional information is meant to aid the solver understand the meaning of a symbol they may not have seen before. We manually encode this information in the feature set associated with each token for synthesis. Where applicable, we also add consonant/vowel distinctions in the given features, since this is a basic distinction assumed in the solutions to many Olympiad problems.

We use the assignments that accompany every problem as the test set, ensuring that the correct answer can be inferred based on the given data.

\subsection{Dataset statistics}
The dataset we present is highly multilingual. The 34 problems contain samples from 38 languages, drawn from across 19 language families. There are 15 morphophonology problems, 7 multilingual problems, 6 stress, and 6 transliteration problems. The set contains 1452 training words with an average of 43 words per problem, and 319 test words with an average of 9 per problem. Each problem has a matrix that has between 7 and 43 rows, with an average of 23. The number of columns ranges from 2 to 6, with most problems having 2.

\begin{table*}[t]
    \centering
    \small
    \begin{tabular}{lcccccccccc}
    \toprule
        \multirow{2}{*}{\textbf{Model}} & \multicolumn{2}{c}{\textbf{All}} & \multicolumn{2}{c}{\textbf{Morphophonology}} & \multicolumn{2}{c}{\textbf{Multilingual}} & \multicolumn{2}{c}{\textbf{Transliteration}} & \multicolumn{1}{c}{\textbf{Stress}} \\
    \cmidrule(lr){2-3}\cmidrule(lr){4-5}\cmidrule(lr){6-7}\cmidrule(lr){8-9}\cmidrule(lr){10-10} \\
         & \textsc{Exact} & \textsc{chrF} & \textsc{Exact} & \textsc{chrF} & \textsc{Exact} & \textsc{chrF} & \textsc{Exact} & \textsc{chrF} & \textsc{Exact} \\ \midrule
         
         \textsc{NoFeature} & 26.8\% & 0.64 & 30.1\% & 0.72 & 42.1\% & 0.59 & 12.0\% & 0.51 & 15.4\% \\
         \textsc{Token} & \textbf{32.7\%} & 0.63 & 37.5\% & 0.68 & \textbf{45.3\%} & 0.60 & 16.4\% &  0.52 & 22.2\% \\
         \textsc{Feature} & 30.9\% & 0.51 & \textbf{38.6\%} & 0.56 & 39.9\% & 0.42 & 9.5\% & 0.49 & \textbf{23.0\%} \\\midrule
         LSTM & 8.2\% & 0.44 & 9.2\% & 0.49 & 5.7\% & 0.45 & 2.1\% & 0.31 & 15.0\% \\
         Transformer & 5.4\% & 0.42 & 2.3\% & 0.39 & 9.2\% & 0.50 & 1.7\% & 0.42 & 12.6\% \\
         WFST & 20.9\% & 0.56 & 16.3\% & 0.47 & 38.7\% & 0.63 & \textbf{29.7\%} & 0.71 & 2.8\% \\
    \bottomrule
    \end{tabular}
    \caption{Metrics for all problems, and for problems of each type. The \textsc{chfF} score for stress problems is not calculated, and not used to determine the overall \textsc{chrF} score.}
    \label{tab:metricresults}
\end{table*}

\section{Experiments}
\subsection{Baselines}
Given that we model our task as string transduction, we compare with the following transduction models used as baselines in shared tasks on G2P conversion \cite{gorman-etal-2020-sigmorphon} and morphological reinflection \cite{vylomova-etal-2020-sigmorphon}.

\noindent\textbf{Neural:} We use LSTM-based sequence-to-sequence models with attention as well as Transformer models as implemented by \citet{neuraltransducer}. For each problem, we train a single neural model that takes the source and target column numbers, and the source word, and predicts the target word.

\noindent\textbf{WFST:} We use models similar to the pair $n$-gram models \cite{novak_minematsu_hirose_2016}, with the implementation similar to that used by \citet{lee-etal-2020-massively}. We train a model for each pair of columns in a problem. For each test example $M_{ij}$, we find the column with the smallest index $j'$ such that $M_{ij'}$ is non-empty and use $M_{ij'}$ as the source string to infer $M_{ij}$.

Additional details of baselines are provided in \cref{sec:baselines}.

\subsection{Program synthesis experiments}
As discussed in \cref{sec:probstructure}, the examples in a problem are in a matrix, and we synthesize programs to transform entries in one column to entries in another. Given a problem matrix $M$, we refer to a program to transform an entry in column $i$ to an entry in column $j$ as $M_{:i} \rightarrow M_{:j}$. 
To obtain token-level examples, we use the Smith-Waterman alignment algorithm \cite{smith1981identification}, which favours contiguous sequences in aligned strings.

We train three variants of our synthesis system with different scores for the \texttt{Is} and \texttt{IsToken} operators. The first one, \textsc{NoFeature}, does not use features, or the \texttt{Is} predicate. The second one, \textsc{Token}, assigns a higher score to \texttt{IsToken} and prefers more specific rules that reference tokens.  The third one, \textsc{Feature}, assigns a higher score to \texttt{Is} and prefers more general rules that reference features instead of tokens. All other aspects of the model remain the same across variants. 

\noindent\textbf{Morphophonology and multilingual problems:} For every pair of columns $(s, t)$ in the problem matrix $M$, we synthesize the program $M_{:s} \rightarrow M_{:t}$. To predict the form of a test sample $M_{ij}$, we find a column $k$ such that the program $M_{:k} \rightarrow M_{:j}$ has the best ranking score, and evaluate it on $M_{ik}$.

\noindent\textbf{Transliteration problems:} Given a problem matrix $M$, we construct a new matrix $M'$ for each pair of columns $(s, t)$ such that all entries in $M'$ are in the same script. We align word pairs $(M_{is}, M_{it})$ using the Phonetisaurus many-to-many alignment tool \cite{jiampojamarn-etal-2007-applying}, and build a simple mapping $f$ for each source token to the target token with which it is most frequently aligned. We fill in $M'_{is}$ by applying $f$ to each token of $M_{is}$ and $M'_{it} = M_{it}$. We then find a program $M'_{:s} \rightarrow M'_{:t}$.

\noindent\textbf{Stress problems:} For these problems, we do not perform any alignment, since the training pairs are already token aligned. The synthesis system learns to transform the source string to the sequence of stress values.

\subsection{Metrics}
We calculate two metrics: exact match accuracy, and \textsc{chrF} score \cite{popovic-2015-chrf}. The exact match accuracy measures the fraction of examples the synthesis system gets fully correct. 
\begin{align*}
    \textsc{Exact} &= \frac{\#\{\text{correctly predicted test samples}\}}{\#\{\text{test samples}\}}
\end{align*}
The \textsc{chrF} score is calculated only at the token level, and measures the $n$-gram overlaps between the predicted answer and the true answer, and allows us to measure partially correct answers. We do not calculate the \textsc{chrF} score for stress problems as $n$-gram overlap is not a meaningful measure of performance for these problems.

\subsection{Results}
\cref{tab:metricresults} summarizes the results of our experiments. We report the average of each metric across problems for all problems and by category.

We find that neural models that don't have specific inductive biases for the kind of tasks we present here are not able to perform well with this amount of data. The synthesis models do better than the WFST baseline overall, and on all types of problems except transliteration. This could be due to the simple map computed from alignments before program synthesis causing errors that the rule learning process cannot correct.

\section{Analysis}
We examine two aspects of the program synthesis models we propose. The first is the way it uses the explicit knowledge in the DSL and implicit knowledge provided as the ranking score to generalize. We then consider specific examples of problems, and show examples of where our models succeed and fail in learning different types of patterns.

\begin{table}[h]
    \small
    \centering
    \begin{tabular}{cccc}
    \toprule
        \textbf{Model} & 
        $100\%$ & 
        $\ge 75\%$  & 
        $\ge 50\%$ \\ \midrule
        \textsc{NoFeature} & 3 & 5 & 7 \\ 
        \textsc{Token} & 3 & 6 & 10 \\ 
        \textsc{Feature} & \textbf{3} & \textbf{6} & \textbf{11} \\
        WFST & 1 & 2 & 7 \\
        \bottomrule
    \end{tabular}
    \caption{Number of problems where the model achieves different thresholds of the \textsc{Exact} score.}
    \label{tab:threshold}
\end{table}

\subsection{Features aid generalization}
Since the test examples are chosen to test specific rules, solving more test examples correctly is indicative of the number of rules inferred correctly. In \cref{tab:threshold}, we see that providing the model with features allows it to infer more general rules, solving a greater fraction of more problems. We see that allowing the model to use features increases its performance, and having it prefer more general rules involving features lets it do even better.

\subsection{Correct programs are short}

\begin{figure}[t]
    \centering
    \includegraphics[width=\columnwidth]{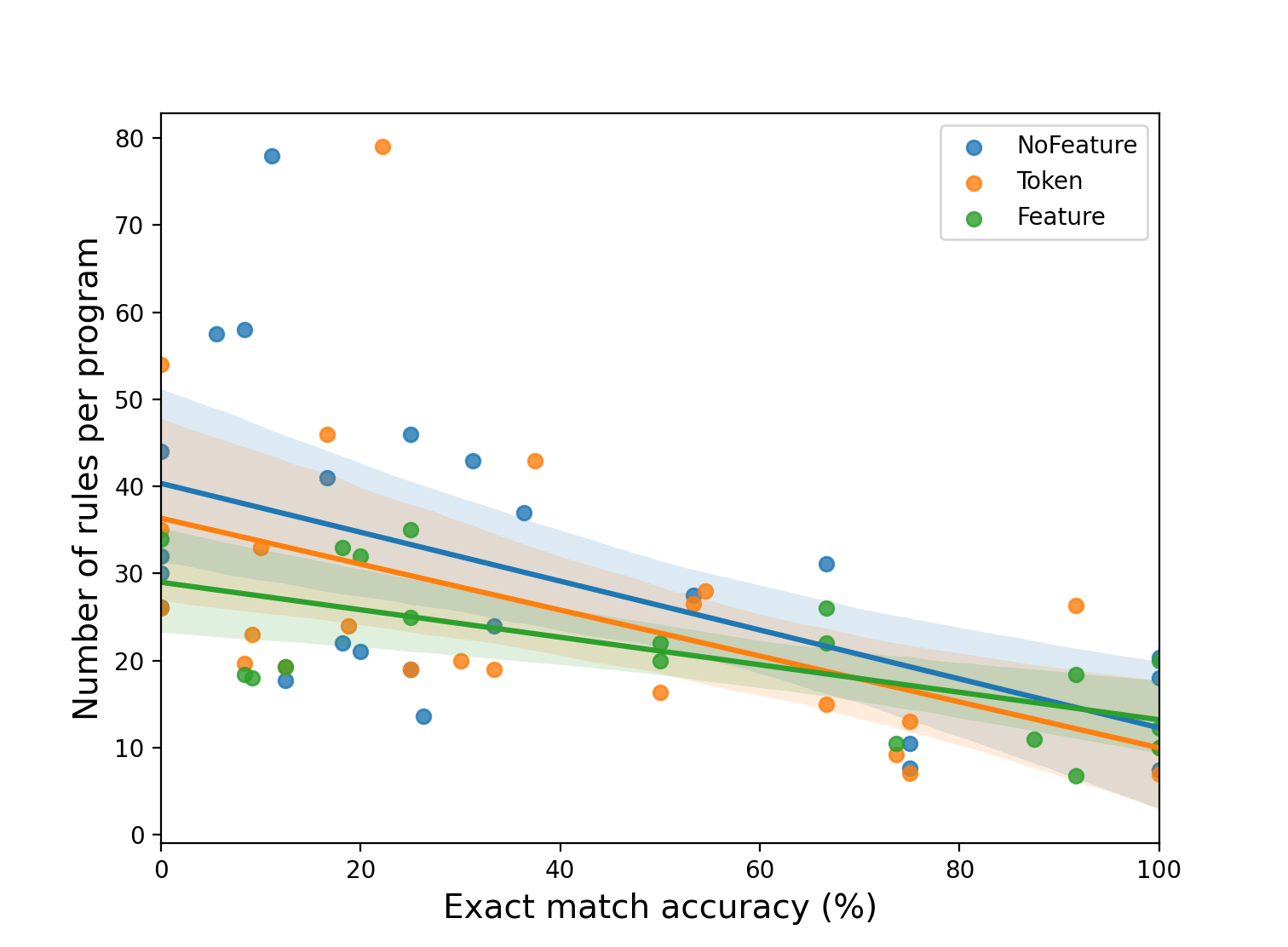}
    \caption{Number of rules plotted against \textsc{Exact} score}
    \label{fig:rulesplot}
\end{figure}

In \cref{fig:rulesplot} we see that the number of rules in a problem\footnote{To account for some problems having more columns than others (and hence more rules), we find the average number of rules for each pair of columns.} tends to be higher when the model gets the problem wrong, than when it gets it right. This indicates that when the model finds many specific rules, it overfits to the training data, and fails to generalize well. This holds true for all the variants,
as seen in the downward slope of the lines.

We also find that allowing and encouraging a model to use features leads to shorter programs. The average length of a program synthesized by \textsc{NoFeatures} is 30.5 rules, while it is 25.8 for \textsc{Token}, and 20.7 for \textsc{Feature}. This suggests that explicit access to features, and implicit preference for them leads to fewer, more general rules.

\subsection{Using features}
Some problems provide additional information about certain sounds. For example, a problem based on the alternation retroflexes in Warlpiri words \cite{warlpiri_problem} explicitly identifies retroflex sounds in the problem statement. In this case, a program produced by our \textsc{Feature} system is able to use these features, and isolate the focus of the problem by learning rules such as 
\begin{figure}[!h]
    \centering
    \vspace{-1cm}
\begin{programs}
    IfThen(Not(Is(w, "retroflex", 0)), 
        Identity(x))
\end{programs}
    \vspace{-0.5cm}
    \label{fig:my_label}
\end{figure}

The system learns a concise solution, and is able to generalize using features rather than learning separate rules for individual sounds.

In the case of inflecting a Mandar verb \cite{mandar_problem}, the \textsc{Feature} system uses a feature to find a more general rule than is the case. To capture the rule that the prefix \textit{\textipa{di-}} changes to \textit{\textipa{mas-}} when the root starts with \textit{\textipa{s}}, the model synthesizes
\begin{figure}[!h]
    \centering
    \vspace{-1cm}
\begin{programs}
    IfThen(Is(w, "fricative", 1),
        ReplaceBy(x, "i", "s"))
\end{programs}
    \vspace{-0.5cm}
    \label{fig:my_label}
\end{figure}

However, since \textit{\textipa{s}} is the only fricative in the data, this rule is equivalent to a rule specific to \textit{\textipa{s}}. This rule also covers examples where the root starts with \textit{\textipa{s}}, and causes the model to miss the more general rule of a voiceless sound at the beginning of the root to be copied to the end of the prefix. It identifies this rule only for roots starting with \textit{\textipa{p}} as
\begin{figure}[!h]
    \centering
    \vspace{-1cm}
\begin{programs}
    IfThen(IsToken(w, "p", 1), 
        CopyReplace(x, w, 1))
\end{programs}
    \vspace{-0.5cm}
    \label{fig:my_label}
\end{figure}

The \textsc{Token} system does not synthesize these rules based on features, and instead chooses rules specific to each initial character in the root.

Since the DSL allows for substituting one token with one other, or inserting multiple tokens, the system has to use multiple rules to substitute one token with multiple tokens. In the case of 
Mandar, we see one way it does this, by performing multiple substitutions (to transform \textit{\textipa{di-}} to \textit{\textipa{mas-}} it replaces \textit{\textipa{d}} and \textit{\textipa{i}} with \textit{\textipa{a}} and \textit{\textipa{s}} respectively, and then inserts \textit{\textipa{m}}). 

\subsection{Multi-pass rules}
In a problem on Somali verb forms \cite{somali_problem}, we see a different way of handling multi-token substitutions by using multi-pass rules to create a complex rule using simpler elements. The problem requires being able to convert verbs from 1st person to 3rd person singular. 
The solution includes a rule where a single token (\textit{\textipa{l}}) is replaced with (\textit{\textipa{sh}}).
The learned program uses two passes to capture this rule through sequential application of two rules: first \texttt{ReplaceBy(x, "l", "h")}, followed by 
    
\begin{figure}[!h]
    \centering
    \vspace{-1cm}
\begin{programs}
    IfThen(TransformationApplied(w, 
            "{ReplaceBy, h}", 1), 
        Insert(x, "s"))
\end{programs}
    \vspace{-0.5cm}
    \label{fig:my_label}
\end{figure}

\subsection{Selecting spans of the input}
In a problem involving reduplication in Tarangan \cite{tarangan_problem}, all variants fail to capture any synthesis rules. Reduplication in Tarangan involves copying one or two syllables in the source word to produce the target word. However, the DSL we use does not have any predicates or transformations that allow the system to reference a span of multiple tokens (which would form a syllable) in the input. Therefore, it fails to 
model reduplication.

\subsection{Global constraints}
Since we provide the synthesis model with token-level examples, 
it does not have access to word-level information.
This results in poor performance on stress problems, as stress depends on the entire word.
Consider the example of Chickasaw stress \cite{chickasaw_problem}. It correctly learns the rule
\begin{figure}[!h]
    \centering
    \vspace{-1cm}
\begin{programs}
    IfThen(Is(w, "long", 0), 
        ReplaceAnyBy(x, "1"))
\end{programs}
    \vspace{-0.5cm}
    \label{fig:my_label}
\end{figure}

that stresses any long vowel in the word. However, since it cannot check if the word has a long vowel that has already been stressed, it is not able to correctly model the case when the word doesn't have a long vowel. 
This results in some samples being marked with stress at two locations, one where the rule for long vowels applies, and one where the rule for words without long vowels applies.

\section{Related work}
\citet{gildea-jurafsky-1996-learning} also study the problem of learning phonological rules from data, and explicitly controlling generalization behaviour. We pursue a similar goal, but 
in a few-shot setting.

\citet{barke-etal-2019-constraint} and \citet{NIPS2015_5785} study program synthesis applied to linguistic rule learning. They make much stronger assumptions about the data (the existence of an underlying form, and the availability of additional information like IPA features). We take a different approach, and study program synthesis models that can work only on the tokens in the word (like \textsc{NoFeature}), and also explore the effect of providing features in these cases. We also test our approach on a more varied set of problems that involves aspects of morphology, transliteration, multilinguality, and stress.

\citet{sahin-etal-2020-puzzling} also present a 
set of Linguistics Olympiad problems as a test of the metalinguistic reasoning abilities of NLP models. While problems in their set involve finding phonological rules, they also require the knowledge of syntax and semantics that are out of the scope of our study. We present a set of problems that only requires reasoning about surface word forms, 
and 
without requiring the
meanings.

\section{Conclusion}
In this paper, we explore the problem of learning linguistic rules from only a few training examples. 
We approach this using program synthesis,
and demonstrate that it is a powerful and flexible technique for 
learning phonology rules in 
Olympiad problems. These problems are designed to be challenging 
tasks that require learning rules from a minimal number of examples. 
These problems also allow us to specifically test for generalization. 

We compare our approach to various baselines, and find that 
it is capable of learning phonological rules that generalize 
much better than existing approaches. We show that using the DSL, we can explicitly control the structure of rules, and using the ranking score, we can provide the model with implicit preferences for certain kinds of rules.

Having demonstrated the potential of program synthesis as a learning technique that can work with very little data and provide human-readable models, we hope to apply it to learning more complex types of lingusitic rules in the future. 

In addition to being a way to learn rules from data, the ability to explicity control the generalization behaviour of the model allows for the use of program synthesis to understand the kinds of learning biases and operations that are required to model various linguistic processes. We leave this exploration to future work.

\section*{Acknowledgements}
We would like to thank Partho Sarthi for invaluable help with \textsc{prose} and \textsf{NDSyn}. We would also like to thank the authors of the ProLinguist paper for their assistance. Finally, we would like to thank the anonymous reviewers for their feedback.

\bibliographystyle{acl_natbib}
\bibliography{anthology,acl2021}

\appendix

\section{\textsf{NDSyn} algorithm}
\label{sec:ndsyn}
We use the \textsf{NDSyn} algorithm to learn disjunctions of rules. We apply \textsf{NDSyn} in multiple passes to allow the model to learn multi-pass rules.

At each pass, the algorithm learns rules to perform token-level transformations that are applied to each element of the input sequence.
The token-level examples are passed to \textsf{NDSyn}, which learns the if-then-else statements that constitute a set of rules. This is done by first generating a set of candidate rules by randomly sampling a token-level example and synthesizing a set of rules that satisfy the example. Then, rules are selected to cover the token-level examples.

Rules that satisfy a randomly sampled example are learnt using the \textsf{FlashMeta} program synthesis algorithm \cite{flashmeta}. The synthesis task is given by the DSL operator $P$ and the specification of constraints $\mathcal{X}$ that the synthesized program must satisfy. In our application, this specification is in the form of token-level examples, and the DSL operators are the predicates and transformations defined in the paper. The algorithm recursively decomposes the synthesis problem $(P, \mathcal{X})$ into smaller tasks $(P_i, \mathcal{X}_i)$ for each argument $P_i$ to the operator. $\mathcal{X}_i$ is inferred using the inverse semantics of the operator $P_i$, which is encoded as a \textit{witness function}. The inverse semantics provides the possible values for the arguments of an operator, given the output of the operator. We refer the reader to the paper by \citet{flashmeta} for a full description of the synthesis algorithm.

After the candidates are generated, they are ranked according to a ranking score of each program. The ranking score for an operator in a program is computed as a function of the scores of its arguments. The arguments may be other operators, offsets, or other constants (like tokens or features). The score for an operator in the argument is computed recursively. The score for an offset favours smaller numbers and local rules by decreasing the score for larger offsets. The score for other constants is chosen to be a small negative constant. The scores for the arguments are added up, along with a small negative penalty to favour shorter programs, to obtain the final score for the operator. 

This ranking score selects for programs that are shorter, and favours either choosing more general by giving the \texttt{Is} predicate a higher score (\textsc{Feature}) or more specific rules by giving the \texttt{IsToken} predicate a higher score (\textsc{Token}).
The top $k$ programs according to the ranking function are chosen as candidates for the next step.

To choose the final set of rules from the candidates generated using the \textsf{FlashMeta} algorithm, we use a \textit{set covering} algorithm that chooses the rules that correctly answer the most number of examples while also incorrectly answering the least. These rules are applied to each example, and the output tokens are tagged with the transformation that is applied. These outputs are then the input to the next pass of the algorithm.

\section{Dataset}
\label{sec:dataappendix}
We select problems from various Linguistics Olympiads to create our dataset. We include publicly available problems that have appeared in Olympiads before. We choose problems that only involve rules based on the symbols in the data, and not based on knowledge of notions such as gender, tense, case, or semantic role. These problems are based on the phonology of a particular language, and include aspects of morphology and orthography, and maybe also the phonology of a different language.  In some cases where a single Olympiad problem involves multiple components that can be solved independent of each other, we include them as separate problems in our dataset.

We put the data and assignments in a matrix, as described in Section~3.1
. We separate tokens in a word by a space while transcribing the problems from their source PDFs. We do not separate diacritics as different tokens, and include them as part of the same token. For each token in the Roman script, we add the boolean features \texttt{vowel} and \texttt{cons}onant, and manually tag the tokens according to whether they are a vowel or consonant.

We store the problems in JSON files with details about the languages, the families to which the languages belong, the data matrix, the notes used to create the features, and the feature sets for each token. 

\section{Baselines}
\label{sec:baselines}

\subsection{Neural}
Following \citet{sahin-etal-2020-puzzling}, we use small neural models for sequence-to-sequence tasks. We train a single neural model for each task, and provide the column numbers as tags in addition to the source sequence. We find that the single model approach works better than training a model for each pair of columns.

\noindent\textbf{LSTM:} We use LSTM models with soft attention \cite{luong2015effective}, with embeddings of size 64, hidden layers of size 128, a 2-layer encoder and a single layer decoder. We apply a dropout of 0.3 for all layers. We train the model for 100 epochs using the Adam optimizer with a learning rate of $10^{-3}$, learning rate reduction on plateau, and a batch size of 2. We clip the gradient norm to 5.

\noindent\textbf{Transformer:} We use Transformer models \cite{vaswani2017attention} with embeddings of size 128, hidden layers of size 256, a 2-layer encoder and a 2-layer decoder. We apply a dropout of 0.3 for all layers. We train the model for 2000 steps using the Adam optimizer with a learning rate of $10^{-3}$, warmup of 400 steps, learning rate reduction on plateau, and a batch size of 2. We use a label smoothing value of 0.1, and clip the gradient norm to 1.

We use the implementations provided at \url{https://github.com/shijie-wu/neural-transducer/} for all neural models.

\subsection{WFST}
We use the implementation the WFST models available at \url{https://github.com/sigmorphon/2020/tree/master/task1/baselines/fst} for the WFST models. We train a model for each pair of columns. We report the results for models of order 5, which were found to perform the best on the test data (highest \textsc{Exact} score) among models of order 3 to 9.

\end{document}